\documentclass[10pt,twocolumn]{article}
\usepackage{amsmath,amssymb,amsfonts}
\usepackage{graphicx}
\usepackage{hyperref}
\usepackage{xcolor}
\usepackage[papersize={216mm,285mm},textheight=234mm,left=21mm, right=21mm, top=27mm, columnsep=5.5mm,footskip=9mm]{geometry}

\begin{document}

\title {Deep Image Synthesis from Intuitive User Input: A Review and Perspectives}

\author{Yuan Xue$^{1}$, Yuan-Chen Guo$^{2}$, Han Zhang$^3$, \\
Tao Xu$^4$,  Song-Hai Zhang$^2$, Xiaolei Huang$^1$ \\
\\
$^{1}$ The Pennsylvania State University, University Park, PA, USA  \\
$^{2}$  Tsinghua University, Beijing, China \\
$^{3}$  Google Brain, Mountain View, CA, USA\\
$^{4}$ Facebook, Menlo Park, CA, USA
}

\date{}

\maketitle

\begin{abstract}
In many applications of computer graphics, art and design, it is desirable for a user to provide intuitive non-image input, such as text, sketch, stroke, graph or layout, and have a computer system automatically generate photo-realistic images that adhere to the input content. While classic works that allow such automatic image content generation have followed a framework of image retrieval and composition, recent advances in deep generative models such as generative adversarial networks (GANs), variational autoencoders (VAEs), and flow-based methods have enabled more powerful and versatile image generation tasks.  This paper reviews recent works for image synthesis given intuitive user input, covering advances in input versatility, image generation methodology, benchmark datasets, and evaluation metrics. This motivates new perspectives on input representation and interactivity, cross pollination between major image generation paradigms, and evaluation and comparison of generation methods.
\\


\end{abstract}

\section{Introduction}\label{sec:introduction}

Machine learning and artificial intelligence have given computers the abilities to mimic or even defeat humans in tasks like playing chess and Go games, recognizing objects from images, translating from one language to another. An interesting next pursuit would be: can computers mimic creative processes such as mimicking painters in making pictures, assisting artists or architects in making artistic or architectural designs?  In fact, in the past decade, we have witnessed advances in systems that synthesize an image from text description \cite{zhang2017stackgan,qiao2019mirrorgan,zhu2019dm, zhang2021cross} or from learned style constant \cite{karras2019style}, 
paint a picture given a sketch \cite{sangkloy_scribbler_2017,ghosh_interactive_2019,gao_sketchycoco_2020,liu2020unsupervised}, render a photorealistic scene from a wireframe \cite{li2019layoutgan,xue2020neural}, create virtual reality content from images and videos \cite{wang2020vr}, among others.  A comprehensive review of such systems can inform about the current state-of-the-art in such pursuits, reveal open challenges and illuminate future directions. In this paper, we make an attempt at a comprehensive review of image synthesis and rendering techniques given simple, intuitive user inputs such as text, sketches or strokes, semantic label maps, poses, visual attributes, graphs and layouts.
We first present ideas on what makes a good paradigm for image synthesis from intuitive user input and review popular  metrics for evaluating the quality of generated images. We then introduce several mainstream methodologies for image synthesis given user inputs, and review algorithms developed for application scenarios specific to different formats of user inputs.  We also summarize major benchmark datasets used by current methods, and advances and trends in image synthesis methodology. Last, we provide our perspective on future directions towards developing image synthesis models capable of generating complex images that are closely aligned with user input condition, have high visual realism, and adhere to constraints of the physical world.


\section{What Makes a Good Paradigm for Image Synthesis from Intuitive User Input?}\label{sec:background}

\subsection{What Types of User Input Do We Need?}
\label{sec:input-characteristics}
For an image synthesis model to be user-friendly and applicable in real-world applications, user inputs that are intuitive, easy for interactive editing, and commonly used in the design and creation processes are desired. We define an input modality to be intuitive if it has the following characteristics:

\begin{itemize}
    \item Accessibility. The input should be easy to access, especially for non-professionals. Take sketch for an example, even people without any trained skills in drawing can express rough ideas through sketching.
    \item Expressiveness. The input should be expressive enough to allow someone to convey not only simple concepts but also complex ideas. 
    \item Interactivity. The input should be interactive to some extent, so that users can modify the input content interactively and fine tune the synthesized output in an iterative fashion.
    \end{itemize}

Taking painting as an example, a sketch is an intuitive input because it is what humans use to design the composition of the painting. On the other hand, being intuitive often means that the information provided by the input is limited, which makes the generation task more challenging. Moreover, for different types of applications, the suitable forms of user input can be quite different.  

For image synthesis with intuitive user input, the most relevant and well-investigated method is with conditional image generation models. In other words, user inputs are treated as conditional input to the synthesis model to guide the generation process by conditional generative models. In this review, we will mainly discuss mainstream conditional image generation applications including those using text descriptions, sketches or strokes, semantic maps, poses, visual attributes, or graphs as intuitive input. The processing and representation of user input are usually application- and modality-dependent. 
When given text descriptions as input, pretrained text embeddings are often used to convert text into a vector-representation of input words. Image-like inputs, such as sketches, semantic maps and poses, are often represented as images and processed accordingly. In particular, one-hot encoding can be used in semantic maps to represent different categories, and keypoint maps can be used to encode poses where each channel represents the position of a body keypoint; both result in multi-channel image-like tensors as input. Using visual attributes as input is most similar to general conditional generation tasks, where attributes can be provided in the form of class vectors. For graph-like user inputs, additional processing steps are required to extract relationship information represented in the graphs. For instance, graph convolutional networks (GCNs)~\cite{kipf2016semi} can be applied to extract node features from input graphs. More details of the processing and representation methods of various input types will be reviewed and discussed in Sec.~\ref{sec:Applications}.

\subsection{How Do We Evaluate the Output Synthesized Images?}
\label{sec:evaluation_metrics}
The goodness of an image synthesis method depends on how well its output adheres to user input, whether the output is photorealistic or structurally coherent, and whether it can generate a diverse pool of images that satisfy requirements.  There have been general metrics designed for evaluating the quality and sometimes diversity of synthesized images. Widely adopted metrics use different methods to extract features from images then calculate different scores or distances. Such metrics include Peak Signal-to-Noise Ratio (PSNR), Inception Score (IS), Fr\'echet inception distance (FID), structural similarity index measure (SSIM) and Learned Perceptual Image Patch Similarity (LPIPS). 

Peak Signal-to-Noise Ratio (PSNR) measures the physical quality of a signal by the ratio between the maximum possible power of the signal and the power of the noise affecting it. For images, PSNR can be represented as
\begin{equation}
\mathrm{PSNR}=\frac{1}{3} \sum_{k} 10 \log _{10} \frac{\max \mathrm{DR}^{2}}{\frac{1}{m} \sum_{i, j}\left(t_{i, j, k}-y_{i, j, k}\right)^{2}}
\end{equation}
where $k$ is the number of channels, DR is the dynamic range of the image ($255$ for 8-bit images), $m$ is the number of pixels, $i,j$ are indices iterating over every pixel, $t$ and $y$ are the reference image and synthesized image respectively.

The Inception Score (IS)~\cite{salimans2016improved} uses a pre-trained Inception~\cite{szegedy2016rethinking} network to compute the KL-divergence between the conditional
class distribution and the marginal class distribution. The inception score is defined as 
\begin{equation}
\begin{split}
\text{IS} = \exp (\mathbb{E}_x \text{KL} (P (y|x) || P (y))),
\end{split}
\end{equation}
where $x$ is an input image and $y$ is the label predicted by an Inception model. A high inception score indicates that the generated images are diverse and semantically meaningful. 

Fr\'echet Inception Distance (FID)~\cite{heusel2017gans} is a popular evaluation metric for image synthesis tasks, especially for Generative Adversarial network (GAN) based models. It computes the divergence between the synthetic data distribution and the real data distribution:
\begin{equation}
\text{FID} = ||\hat{m} - m||_2^2 + \text{Tr}(\hat{C} + C -2(C\hat{C})^{1/2}),
\end{equation}
where $m, C$ and $\hat{m}, \hat{C}$ represent the mean and covariance of the feature embeddings of the real and the synthetic distributions, respectively. The feature embedding is extracted from a pre-trained Inception-v3~\cite{szegedy2016rethinking} model. 

Structural Similarity Index Measure (SSIM)~\cite{wang2004image} or multi-scale structural similarity (MS-SSIM) metric~\cite{wang2003multiscale} gives a relative similarity score to an image against a reference one, which is different from absolute measures like PSNR. The SSIM is defined as:
\begin{equation}
\operatorname{SSIM}(x, y)=\frac{\left(2 \mu_{x} \mu_{y}+c_{1}\right)\left(2 \sigma_{x y}+c_{2}\right)}{\left(\mu_{x}^{2}+\mu_{y}^{2}+c_{1}\right)\left(\sigma_{x}^{2}+\sigma_{y}^{2}+c_{2}\right)}\enspace,
\end{equation}
where $\mu$ and $\sigma$ indicate the average and variance of two windows $x$ and $y$, $c_1$ and $c_2$ are two variables to stabilize the division with weak denominator.
The SSIM measures perceived image quality considering structural information. It tests pair-wise similarity between generated images, where a lower score indicates higher diversity of generated images (i.e. less mode collapses).

Another metric based on features extracted from pre-trained CNN networks is the Learned Perceptual Image Patch Similarity (LPIPS) score~\cite{zhang2018unreasonable}. The distance is calculated as 
\begin{equation}
d\left(x, x_{0}\right)=\sum_{l} \frac{1}{H_{l} W_{l}} \sum_{h, w}\left\|w_{l} \odot\left(\hat{y}_{h w}^{l}-\hat{y}_{0 h w}^{l}\right)\right\|_{2}^{2}\enspace,
\end{equation}
where $\hat{y}^{l}, \hat{y}_{0}^{l} \in \mathbb{R}^{H_{l} \times W_{l} \times C_{l}}$ are unit-normalized feature stack from the $l$-th layer in a pre-trained CNN and $w_{l}$ indicates channel-wise weights.
LPIPS evaluates perceptual similarity between image patches using the learned deep features from trained neural networks.

For flow based models \cite{rezende2015variational, kingma2018glow} and autoregressive models \cite{van2016pixel, van2016conditional, salimans2017pixelcnn++}, the average negative log-likelihood (\textit{i.e.}, bits per dimension)~\cite{van2016pixel} is often used to evaluate the quality of generated images. It is calculated as the negative log-likelihood with log base 2 divided by the number of pixels, which is interpretable as the
number of bits that a compression scheme based on this
model would need to compress every RGB color value~\cite{van2016pixel}.

Except for metrics designed for general purposes, specific evaluation metrics have been proposed for different applications with various input types. For instance, using text descriptions as input,
R-precision~\cite{xu2018attngan} evaluates whether a generated image is well conditioned on the
given text description. The R-precision is measured by retrieving relevant text given an image query. For sketch-based image synthesis, classification accuracy is used to measure the realism of the synthesized objects \cite{ghosh_interactive_2019,gao_sketchycoco_2020} and how well the identities of synthesized results match those of real images \cite{lu_image_2018}. Also, similarity between input sketches and edges of synthesized images can be measured to evaluate the correspondence between the input and output \cite{gao_sketchycoco_2020}. In the scenario of pose-guided person image synthesis, ``masked'' versions of IS and SSIM, Mask-IS and Mask-SSIM are often used to ignore the effects of background \cite{ma_pose_2018,ma_disentangled_2018,siarohin_deformable_2018,song2019unsupervised,zhu2019progressive}, since we only want to focus on the synthesized human body. Similar to sketch-based synthesis, detection score (DS) is used to evaluate how well the synthesized person can be detected \cite{siarohin_deformable_2018,zhu2019progressive} and keypoint accuracy can be used to measure the level of correspondence between keypoints \cite{zhu2019progressive}. For semantic maps, a commonly used metric tries to restore the semantic-map input from generated images using a pre-trained segmentation network and then compares the restored semantic map with the original input by Intersection over Union (IoU) score or other segmentation accuracy measures.
Similarly, using visual attributes as input,  
a pre-trained attribute classifier or regressor can be used to assess the attribute correctness of generated images.


\section{Overview of Mainstream Conditional Image Synthesis Paradigms} \label{sec:overview}
Image synthesis models with intuitive user inputs often involve different types of generative models, more specifically, conditional generative models that treat user input as observed conditioning variable.
Two major goals of the synthesis process are high realism of the synthesized images, and correct correspondences between input conditions and output images. In existing literature, methods vary from more traditional retrieval and composition based methods to more recent deep learning based algorithms. In this section, we give an overview of the architectures and main components of different conditional image synthesis models. 

\subsection{Retrieval and Composition}
Traditional image synthesis techniques mainly take a retrieval and composition paradigm. In the retrieval stage, candidate images / image fragments are fetched from a large image collection, under some user-provided constraints, like texts, sketches and semantic label maps. Methods like edge extraction, saliency detection, object detection and semantic segmentation are used to pre-process images in the collection according to different input modalities and generation purposes, after which the retrieval can be performed using shallow image features like HoG and Shape Context \cite{belongie2001shape}. The user may interact with the system to improve the quality of the retrieved candidates. In the composition stage, the selected images or image fragments are combined by Poisson Blending, Alpha blending, or a hybrid of both \cite{chen_sketch2photo_2009}, resulting in the final output image. 

The biggest advantage of synthesizing images through retrieval and composition is its controllability and interpretability. The user can simply intervene with the generation process in any stage, and easily find out whether the output image looks like the way it should be.  But it can not generate instances that do not appear in the collection, which restricts the range and diversity of the output.

\subsection{Conditional Generative Adversarial Networks (cGANs)}
Generative Adversarial Networks (GANs)~\cite{goodfellow2014generative} have achieved tremendous success in various image generation tasks. A GAN model typically consists of two networks: a generator network that learns to generate realistic synthetic images and a discriminator network that learns to differentiate between real images and synthetic images generated by the generator. The two networks are optimized alternatively through adversarial training. Vanilla GAN models are designed for unconditional image generation, which implicitly model the distribution of images.  To gain more control over the generation process, conditional GANs or cGANs~\cite{mirza2014conditional} synthesize images based on both a random noise vector and a condition vector provided by users. The objective of training cGAN as a minimax game is
\begin{equation}
\begin{split}
\min_{\theta_G} \max_{\theta_D} & \mathcal{L}_{\text{cGAN}} =\  \mathbb{E}_{(x, y)\sim p_\mathrm{data}(x, y)}[\log D(x, y)] \ +\\& \mathbb{E}_{z\sim p(z), y \sim p_\mathrm{data}(y)}[\log (1 - D(G(z, y), y)] \enspace ,\label{Eq:cGAN}
\end{split}
\end{equation}
where $x$ is the real image, $y$ is the user input, and $z$ is the random noise vector. There are different ways of incorporating user input in the discriminator, such as inserting it at the beginning of the discriminator~\cite{mirza2014conditional}, middle of the discriminator~\cite{miyato2018cgans}, or the end of the discriminator~\cite{odena2017conditional}.

\subsection{Variational Auto-encoders (VAEs)}
\begin{figure}[tbp]
    \centering
    \includegraphics[width=0.95\linewidth]{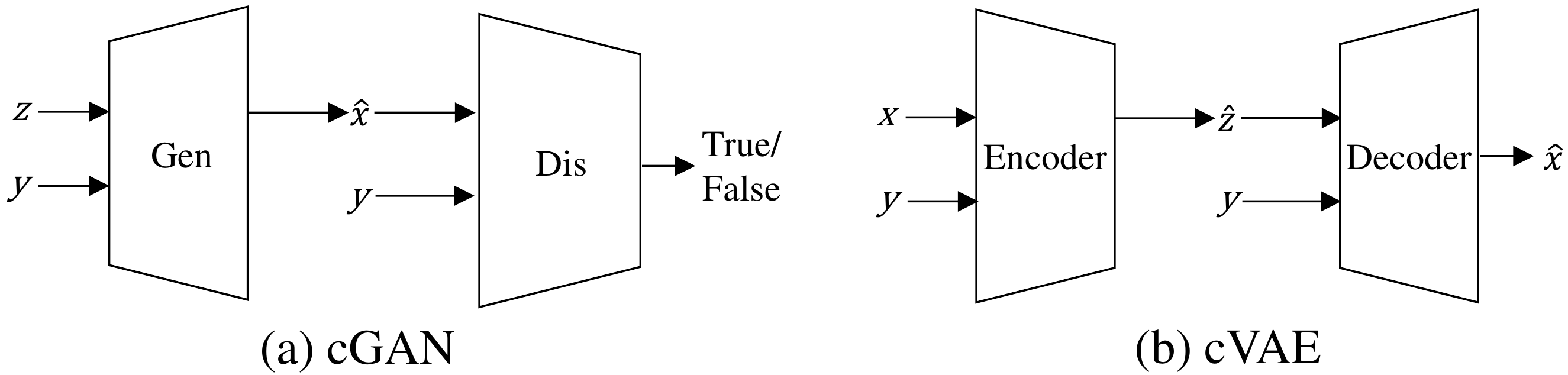}
    \caption{A general illustration of cGAN and cVAE that can be applied to image synthesis with intuitive user inputs. During inference, the generator in cGAN and the decoder in cVAE generate new images $\hat{x}$ under the guidance of user input $y$ and noise vector or latent variable $z$.}
    \label{fig:methods}
\end{figure}

Variational auto-encoders (VAEs) proposed in~\cite{kingma2013auto} extend the idea of auto-encoder and introduce variational inference to approximate the latent representation $z$ encoded from the input data $x$. The encoder converts $x$ into $z$ in a latent space where the decoder tries to reconstruct $x$ from $z$. Similar to GANs which typically assume the input noise vector follows a Gaussian distribution, VAEs use variational inference to approximate the posterior $p(z|x)$ given that $p(z)$ follows a Gaussian distribution.
After the training of VAE, the decoder is used as a generator, similar to the generator in GAN, which can draw samples from the latent space and generate new synthetic data. Based on the vanilla VAE, Sohn~\textit{et al.} proposed a conditional VAE (cVAE)~\cite{sohn2015learning, klys2018learning, ivanov2018variational} which is a conditional
directed graphical model whose input observations modulate the latent variables that generate the outputs. Similar to cGANs, cVAEs allow the user to provide guidance to the image synthesis process via user input. The training objective for cVAE is 
\begin{equation}
\begin{split}
\max_{\theta,\phi} \mathcal{L}_{\text{cVAE}} = \  &\mathbb{E}_{z\sim Q_{\phi}}[\log P_{\theta}(x \mid z, y)]\  -
\\&D_{\text{KL}}[Q_{\phi}(z \mid x, y) \| p(z \mid y)]\enspace,\label{Eq:cVAE}
\end{split}
\end{equation}
where $x$ is the real image, $y$ is the user input, $z$ is the latent variable and $p(z\mid x)$ is the prior distribution of the latent vectors such as the Gaussian distribution. $\phi$ and $\theta$ are parameters of the encoder $Q$ and decoder $P$ networks, respectively. An illustration of cGAN and cVAE can be found in Fig.~\ref{fig:methods}.

\subsection{Other Learning-based Methods}
Other learning-based conditional image synthesis models include hybrid methods such as the combination of VAE and GAN models~\cite{larsen2016autoencoding, bao2017cvae}, autoregressive models and normalizing flow-based models. Among these methods, autoregressive models such as PixelRNN~\cite{van2016pixel}, PixelCNN~\cite{van2016conditional}, and PixelCNN++~\cite{salimans2017pixelcnn++} provide
tractable likelihood over priors such as class conditions. The generation process is similar to an autoregression model: while classic autoregression models predict future information based on past observations, image autoregressive models synthesize next image pixels based on previously generated or existing nearby pixels. 

Flow-based models \cite{rezende2015variational}, or normalizing flow based methods, consist of a sequence of invertible transformations which can convert a simple distribution (e.g., Gaussian) into a more complex one with the same dimension.  While flow based methods have not been widely applied to image synthesis with intuitive user inputs, few works \cite{kingma2018glow} show that they have great potential in visual attributes guided synthesis and may be applicable to broader scenarios.

Among the aforementioned mainstream paradigms, traditional retrieval and composition methods have the advantage of better controllability and interpretability, although the diversity of synthesized images and the flexibility of the models are limited. In comparison, deep learning based methods generally have stronger feature representation capacity, with GANs having the potential of generating images with highest quality. While having been successfully applied to various image synthesis tasks due to their flexibility, GAN models lack tractable and explicit likelihood estimation. On the contrary, autoregressive models admit a tractable likelihood estimation, and can assign a probability to a single sample. VAEs with latent representation learning provide better feature representation power and can be more interpretable. Compared with VAEs and autoregressive models, normalizing flow methods provide both feature representation power and tractable likelihood estimation.  

\section{Methods Specific to Applications with Various Input Types}\label{sec:Applications}


In this section, we review works in the literature that target application scenarios with specific input types. We will review methods for image synthesis from text descriptions, sketches and strokes, semantic label maps, poses, and other input modalities including visual attributes, graphs and layouts. Among the different input types, text descriptions are flexible, expressive and user-friendly, yet the comprehension of input content and responding to interactive editing can be challenging to the generative models; example applications of text-to-image systems are computer generated art, image editing, computer-aided design, interactive story telling and visual chat for education and language learning.  Image-like inputs such as sketches and semantic maps contain richer information and can better guide the synthesis process, but may require more efforts from users to provide adequate input; such inputs can be used in applications such as image and photo editing, computer-assisted painting and rendering. Other inputs such as visual attributes, graphs and layouts allow appearance, structural or other constraints to be given as conditional input and can help guide the generation of images that preserve the visual properties of objects and geometric relations between objects; they can be used in various computer-aided design applications for architecture, manufacturing, publishing, arts, and fashion. 

\subsection{Text Description as Input}
\begin{figure}[tbp]
    \centering
    \includegraphics[width=0.95\linewidth]{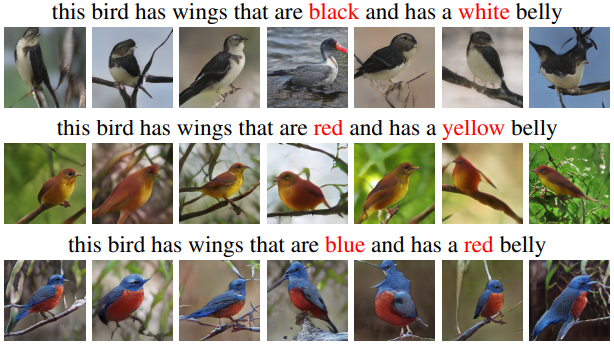}
    \caption{Example bird image synthesis results given text descriptions as input with an attention mechanism. Key words in the input sentences are correctly captured and represented in the generated images. Image taken from AttnGAN ~\cite{xu2018attngan}.}
    \label{fig:attngan}
\end{figure}

The task of text-to-image synthesis (Fig.~\ref{fig:attngan}) is using descriptive sentences as inputs to guide the generation of corresponding images. The generated image types vary from single-object images~\cite{nilsback2008automated, WelinderEtal2010} to multi-object images with complex background ~\cite{lin2014microsoft}. Descriptive sentences in a  natural language offer a general and flexible way of describing visual concepts and objects. As text is one of the most intuitive types of user input, text-to-image synthesis has gained much attention from the research community and numerous efforts have been made towards developing better text-to-image synthesis models. In this subsection, we will review state-of-the-art text-to-image synthesis models and discuss recent advances.

\noindent {\bf Learning Correspondence Between Text and Image Representations.} One of the major challenges of the text-to-image synthesis task is that the input text and output image are in different modalities, which requires learning of correspondence between text and image representations. Such multi-modality nature  and the need to learn text-to-image correspondence motivated Reed~\textit{et al.}~\cite{reed2016generative} to first propose to solve the task using a GAN model. In~\cite{reed2016generative}, the authors proposed to generate images conditioned on the embedding of text descriptions, instead of class labels as in traditional cGANs~\cite{mirza2014conditional}. To learn the text embedding from input sentences, a deep convolutional image encoder and a character level convolutional-recurrent text encoder are trained jointly so that the text encoder can learn a vector-representation of the input text descriptions. Adapted from the DCGAN architecture~\cite{radford2015unsupervised}, the learned text encoding is then concatenated with both the input noise vector in the generator and the image features in the discriminator along the depth dimension. The method \cite{reed2016generative} generated encouraging results on both the Oxford-102 dataset~\cite{nilsback2008automated} and the CUB dataset~\cite{WelinderEtal2010}, with the limitation that the resolution of generated images is relatively low ($64\times64$).
Another work proposed around the same time as DCGAN is by Mansimov~\textit{et al.}~\cite{mansimov2015generating}, which proposes a combination of a recurrent variational
autoencoder with an attention model which iteratively draws patches on a canvas, while attending to the relevant words in the description. Input text descriptions are represented as a sequence of consecutive words and images are represented as a sequence of patches drawn on a canvas. For image generation which samples from a Gaussian distribution, the Gaussian mean and variance depend on the previous hidden states of the generative LSTM. Experiments by~\cite{mansimov2015generating} on the MS-COCO dataset show reasonable results that correspond well to text descriptions.

To further improve the visual quality and realism of generated images given text descriptions, Han~\textit{et al.} proposed multi-stage GAN models, StackGAN~\cite{zhang2017stackgan} and StackGAN++~\cite{zhang2018stackgan++}, to enable multi-scale, incremental refinement in the image generation process. Given text descriptions, StackGAN~\cite{zhang2017stackgan} decomposes the text-to-image
generative process into two stages, where in Stage-I it captures basic object features and background layout, then in Stage-II it refines details of the objects and generates a higher resolution image. Unlike~\cite{reed2016generative} which transforms high dimensional text encoding into low dimensional latent variables, StackGAN adopts a Conditioning Augmentation which is to sample the latent variables from an independent Gaussian
distribution parameterized by the text encoding. Experiments on the Oxford-102~\cite{nilsback2008automated}, CUB~\cite{WelinderEtal2010} and COCO~\cite{lin2014microsoft} datasets show that StackGAN can generate compelling images with resolution up to $256\times256$. In StackGAN++~\cite{zhang2018stackgan++}, the authors extended the original StackGAN into a more general and robust model which contains multiple generators and discriminators to handle images at different resolutions.
Then, Zhang~\textit{et al.}~\cite{zhang2018photographic}  extended the multi-stage generation idea by proposing a HDGAN model with a single-stream generator and multiple hierarchically-nested discriminators for high-resolution image synthesis. Hierarchically-nested discriminators distinguish outputs from intermediate layers of the generator to capture hierarchical visual features. The training of HDGAN is done via optimizing a pair loss~\cite{reed2016generative} and a patch-level discriminator loss~\cite{isola2017image}.

 In addition to generation via multi-stage refinement~\cite{zhang2017stackgan, zhang2018stackgan++}, the attention mechanism is introduced to improve text to image synthesis at a more fine-grained level. Xu~\textit{et al.} introduced AttnGAN~\cite{xu2018attngan}, an attention driven image synthesis model that generates images by focusing on different regions described by different words of the text input. A Deep Attentional Multimodal Similarity Model (DAMSM) module is also proposed to match the learned embedding between image regions and text at the word level.
To achieve better semantic consistency between text and image, Qiao~\textit{et al.}~\cite{qiao2019mirrorgan} proposed MirrorGAN which guides the image generation with both sentence- and word-level attention and further tried to reconstruct the original text input to guarantee the image-text consistency. The backbone of MirrorGAN uses a multi-scale generator as in~\cite{zhang2018stackgan++}. The proposed text reconstruction model is pre-trained to stabilize the training of MirrorGAN.
Zhu~\textit{et al.}~\cite{zhu2019dm} introduces a gating mechanism where a writing gate writes selected important textual features from the given sentence into a dynamic memory, and a response gate adaptively reads from the memory and the visual features from some initially generated images. The proposed DM-GAN relies less on the quality of the initial images and can  refine  poorly-generated initial images with wrong colors and rough shapes.

To learn expression variants in different text descriptions of the same image, 
Yin~\textit{et al.} proposes SD-GAN~\cite{yin2019semantics} to distill the shared semantics from texts that describe the same image. 
The authors propose a Siamese structure with a contrastive loss to minimize the distance between 
images generated from descriptions of the same image, and maximize the distance between those generated from the descriptions of different images. To retain the semantic
diversity for fine-grained image generation, a semantic-conditioned batch normalization is also introduced for enhanced visual-semantic embedding.

\noindent {\bf Location and Layout Aware Generation.}\label{sec:text-layout}
With advances in correspondence learning between text and image, content described in the input text can already be well captured in the generated image. However, to achieve finer control of generated images such as object locations, additional inputs or intermediate steps are often required. For text-based and location-controllable synthesis, Reed~\textit{et al.}~\cite{reed2016learning} proposes to generate images conditioned on both the text description and object locations. Built upon the similar idea of inferring scene structure for image generation, Hong~\textit{et al.}~\cite{hong2018inferring} introduces a novel hierarchical approach for text-to-image synthesis by inferring semantic layout from the text description. Bounding boxes are first generated from text input through an auto-regressive model, then semantic layouts are refined from the generated bounding boxes using a convolutional recurrent neural network. Conditional on both the text and the semantic layouts, the authors adopt a combination of pix2pix~\cite{isola2017image} and CRN~\cite{chen2017photographic} image-to-image translation model to generate the final images. With predicted semantic layouts, this work \cite{hong2018inferring} has potential in generating more realistic images containing complex objects such as those in the MS-COCO~\cite{lin2014microsoft} dataset.  
Li~\textit{et al.}~\cite{li2019object} extends the work by~\cite{hong2018inferring} and introduces Obj-GAN, which generates salient objects given text description. Semantic layout is first generated as in ~\cite{hong2018inferring} then later converted into the synthetic image. A Fast R-CNN~\cite{girshick2015fast} based object-wise discriminator is developed to retain the matching between generated objects and the input text and layout. Experiments on the MS-COCO dataset show improved performance in generating complex scenes compared to previous methods.

Compared to~\cite{hong2018inferring}, Johnson~\textit{et al.}~\cite{johnson2018image} includes another intermediate step which converts the input sentences into scene graphs before generating the semantic layouts. A graph convolutional network is developed to generate embedding vectors for each object. Bounding boxes and segmentation
masks for each object, constituting the scene layout, are converted from the object embedding vectors. Final images are synthesized by a CRN model~\cite{chen2017photographic} from the noise vectors and scene layouts. In addition to text input,~\cite{johnson2018image} also allows direct generation from input scene graphs. Experiments are conducted on Visual Genome~\cite{krishna2017visual} dataset and COCO-Stuff~\cite{caesar2018coco} dataset which is augmented on a subset of the MS-COCO~\cite{lin2014microsoft} dataset, and show better depiction of complex sentences with many objects than previous method~\cite{zhang2017stackgan}.

Without taking the complete semantic layout as additional input, Hinz~\textit{et al.}~\cite{hinz2019generating} introduces a model consisting of a global pathway and an object
pathway for finer control of object location
and size within an image. The global pathway is responsible for creating a general layout of the global scene, while the object pathway generates object features within the given bounding boxes. Then the outputs of the global and object pathways are combined to generate the final synthetic image. When there is no text description available, \cite{hinz2019generating} can take a noise vector and the individual object bounding boxes as input.

Taking an approach different from GAN based methods, Tan~\textit{et al.}~\cite{tan2018text2scene} proposes a Text2Scene model for text-to-scene generation, which learns to sequentially generate objects and their attributes such as location, size, and appearance at every time step. With a convolutional recurrent module and attention module, Text2Scene can generate abstract scenes and object layouts directly from descriptive sentences. For image synthesis, Text2Scene retrieves patches from real images to generate the image composites.

\noindent {\bf Fusion of Conditional and Unconditional Generation.}
While most existing text-to-image synthesis models are based on conditional image generation, Bodla~\textit{et al.}~\cite{bodla2018semi} proposes a FusedGAN which combines unconditional image generation and conditional image generation. An unconditional generator produces a structure prior independent of the condition, and the other conditional generator refines details and creates an image that matches the input condition. FusedGAN is evaluated on both the text-to-image generation task and the attribute-to-face generation task which will be discussed later in Sec.~\ref{app:attribute}. 

\noindent {\bf Evaluation Metrics for Text to Image Synthesis. }
Widely used metrics for image synthesis such as IS~\cite{salimans2016improved} lack awareness of matching between the text and generated images. Recently, more efforts have been focused on proposing more accurate evaluation metrics for text to image synthesis and for evaluating the correspondence between generated image content and input condition. R-precision is proposed in~\cite{xu2018attngan} to evaluate whether a generated image is well conditioned on the given text description. Hinz~\textit{et al.} proposes the Semantic Object Accuracy (SOA) score~\cite{hinz2019semantic} which uses a pre-trained object detector to check whether the generated image contains the objects described in the caption, especially for the MS-COCO dataset. SOA shows better correlation with human perception than IS in the user study and provides a better guidance for training text to image synthesis models.

\noindent {\bf Benchmark Datasets.} For text-guided image synthesis tasks, popular benchmark datasets include datasets with a single object category and datasets with multiple object categories. For single object category datasets, the Oxford-102 dataset~\cite{nilsback2008automated} contains 102 different types of flowers common in the UK. The CUB dataset~\cite{WelinderEtal2010} contains photos of 200 bird species of which mostly are from North America. Datasets with multiple object categories and complex relationships can be used to train models for more challenging image synthesis tasks. One such dataset is MS-COCO~\cite{lin2014microsoft}, which has a training set with 80k images and a validation set with 40k images. Each image in the COCO dataset has five text descriptions.

\subsection{Image-like Inputs}
In this section, we summarize image synthesis works based on three types of intuitive inputs, namely sketch, semantic map and pose. We call them ``image-like inputs'' because all of them can be, and have been represented as rasterized images. Therefore, synthesizing images from these image-like inputs can be regarded as an image-to-image translation problem. Several works provide general solutions to this problem, like pix2pix~\cite{isola2017image} and pix2pixHD~\cite{wang2018high}. In this survey, we focus on works that deal with a specific type of input. 

\vspace{5pt}
\subsubsection{Sketches and Strokes as Input}
Sketches, or line drawings, can be used to express users' intention in an intuitive way, even for those without professional drawing skills. With the widespread use of touch screens, it has become very easy to create sketches; and the research community is paying increasingly more attention to the understanding and processing of hand-drawn sketches, especially in applications such as sketch-based image retrieval and sketch-to-image generation. Generating realistic images from sketches is not a trivial task, since the synthesized images need to be aligned spatially with the given sketches, while maintain semantic coherence.

\begin{figure}[ht]
    \centering
    \includegraphics[width=0.99\linewidth]{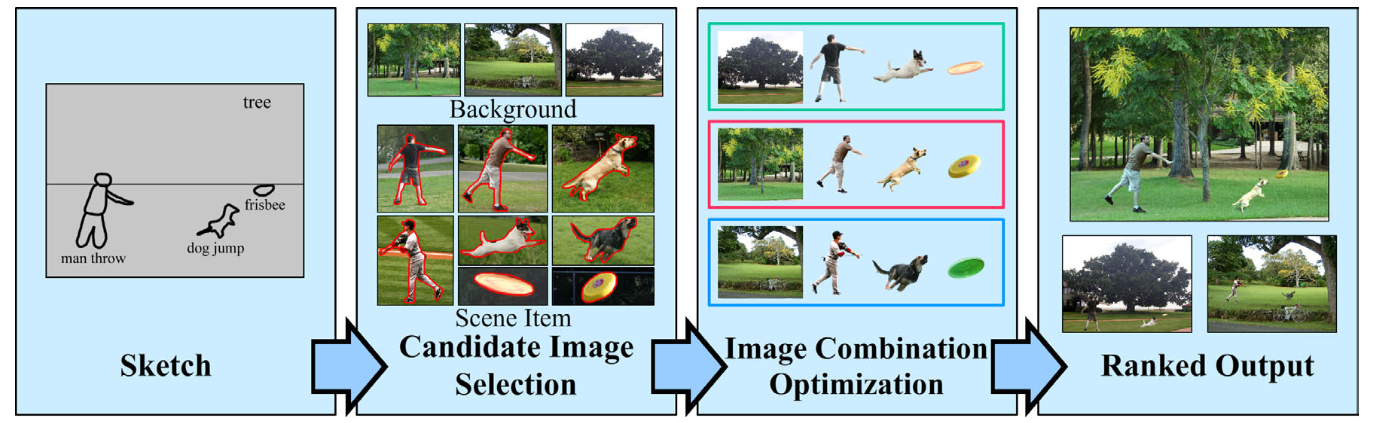}
    \caption{A classical pipeline of retrieval-and-composition methods for synthesis. Candidate images are generated by composing image segments retrieved from a pre-built image database. Image taken from~\cite{chen_sketch2photo_2009}.}
    \label{fig:sketch2photo}
\end{figure}

\noindent {\bf Retrieval-and-Composition based Approaches.}
Early approaches of generating image from sketch mainly take a retrieval-and-composition strategy. For each object in the user-given sketch, they search for candidate images in a pre-built object-level image (fragment) database, using some similarity metric to evaluate how well the sketch matches the image. The final image is synthesized as the composition of retrieved results, mainly by image blending algorithms. Chen et al.~\cite{chen_sketch2photo_2009} presented a system called Sketch2Photo, which composes a realistic image from a simple free-hand sketch annotated with text labels. The authors proposed a contour-based filtering scheme to search for appropriate photographs according to the given sketch and text labels, and
proposed a novel hybrid blending algorithm, which is a combination of alpha blending and Poisson blending, to improve the synthesis quality.
Eitz et al.~\cite{eitz_photosketcher_2011} created Photosketcher, a system that finds semantically relevant regions from appropriate images in a large image collection and composes the regions automatically. Users can also interact with the system by drawing scribbles on the retrieved images to improve region segmentation quality, re-sketching to find better candidates, or choosing from different blending strategies.
Hu et al.~\cite{hu2013patchnet} introduced PatchNet, a hierarchical representation of image regions that summarizes a homogeneous image patch by a graph node and represents geometric relationships between regions by labeled graph edges. PatchNet was shown to be a compact representation that can be used efficiently for sketch-based, library-driven, interactive image editing.
Wang et al.~\cite{wang_mindcamera_2018} proposed a sketch-based image synthesis method that compares sketches with contours of object regions via the GF-HOG descriptor, and novel images are composited by GrabCut followed by Possion blending or alpha blending.
For generating images of a single object like an animal under user-specified poses and appearances, Turmukhambetov et al.~\cite{turmukhambetov_interactive_2015} presented a sketch-based interactive system that generates the target image by composing patches of nearest neighbour images on the joint manifold of ellipses and contours for object parts.

\noindent {\bf Deep Learning based Approaches.}
In recent years, deep convolutional neural networks (CNNs) have achieved significant progress in image-related tasks. CNNs have been used to map sketches to images with the benefit of being able to synthesize novel images that are different from those in pre-built databases.
One challenge to using deep CNNs is that training of such networks require paired sketch-image data, which can be expensive to acquire.
Hence, various techniques have been proposed to generate synthetic sketches from images, and then use the synthetic sketch and image pairs for training. Methods for synthetic sketch generation include boundary detection algorithms such as Canny, Holistically-nested Edge Detection (HED)~\cite{xie2015holistically}, and stylization algorithms for image-to-sketch conversion~\cite{winnemoller2012xdog,kang2007coherent,li2019im2pencil,li2019photo,gastal2011domain}. Post-processing steps are adopted for small stroke removal, spline fitting~\cite{hahn2014autotrace} and stroke simplification~\cite{simo2016learning}.  A few works utilize crowd-sourced free-hand sketches for training~\cite{gao_sketchycoco_2020, liu2020unsupervised}. They either construct pseudo-paired data by matching sketches and images~\cite{gao_sketchycoco_2020}, or propose a method that does not require paired data~\cite{liu2020unsupervised}. Another aspect of CNN training that has been investigated is the representation of sketches. In some works~\cite{chen_sketchygan_2018,li2019linestofacephoto}, the input sketches are transformed into distance fields to obtain a dense representation, but no experimental comparisons have been done to demonstrate which form of input is more suitable for CNNs to process.
Next, we review specific works that utilize a deep-learning based approach for sketch to image generation. 

Treating a sketch as an ``image-like'' input, several works use a fully convolutional neural network architecture to generate photorealistic images.  Gucluturk et al.~\cite{gucluturk_convolutional_2016} first attempted to use deep neural networks to tackle the problem of sketch-based synthesizing. They developed three different models to generate face images from three different types of sketches, namely line sketch, grayscale sketch and color sketch. An encoder-decoder fully convolutional neural network is adopted and trained with various loss terms. A total variation loss is proposed to encourage smoothness.
Sangkloy et al.~\cite{sangkloy_scribbler_2017} proposed Scribbler, a system that can generate realistic images from human sketches and color strokes. XDoG filter is used for boundary detection to generate image-sketch pairs and color strokes are sampled to provide color constraints in training. The authors also use an encoder-decoder network architecture and adopt similar loss functions as in~\cite{gucluturk_convolutional_2016}. The users can interact with the system in real time. The authors also provide applications for colorization of grayscale images.

Generative Adversarial Networks have also been used for sketch-to-image synthesis.  Chen et al.~\cite{chen_sketchygan_2018} proposed a novel GAN-based architecture with multi-scale inputs for the problem. The generator and discriminator both consist of several Masked Residual Unit (MRU) blocks. MRU takes in a feature map and an image, and outputs a new feature map, which can allow a network to repeatedly condition on an input image, like the recurrent network. They also adopt a novel data augmentation technique, which generates sketch-image pairs automatically through edge detection and some post-processing steps including binarization, thinning, small component removal, erosion, and spur removal. To encourage diversity of generated images, the authors proposed a diversity loss, which maximizes the L1 distance between the outputs of two identical input sketches with different noise vectors. 
Lu et al.~\cite{lu_image_2018} considered the sketch-to-image synthesis problem as an image completion task and proposed a contextual GAN for the task. Unlike a traditional image completion task where only part of an object is masked, the entire real image is treated as the missing piece in a joint image that consists of both sketch and the corresponding photo.
The advantage of using such a joint representation is that, instead of using the sketch as a hard constraint, the sketch part of the joint image serves as a weak contextual constraint. Furthermore, the same framework can also be used for image-to-sketch generation where the sketch would be the masked or missing piece to be completed. 
Ghosh et al.~\cite{ghosh_interactive_2019} presents an interactive GAN-based sketch-to-image translation system. As the user draws a sketch of a desired object type, the system automatically recommends completions and fills the shape with class-conditioned texture. 
The result changes as the user adds or removes strokes over time, which enables a feedback loop that the user can leverage for interactive editing.  The system consists of a shape completion stage based on a non-image generation network~\cite{mescheder2018training}, and a class-conditioned appearance translation stage based on the encoder-decoder model from MUNIT~\cite{huang2018multimodal}. To perform class-conditioning more effectively, the authors propose a soft gating mechanism, instead of using simple concatenation of class codes and features.

Several works focus on sketch-based synthesis for human face images. Portenier et al.~\cite{portenier2018faceshop} developed an interactive system for face photo editing. The user can provide shape and color constraints by sketching on the original photo, to get an edited version of it. The editing process is done by a CNN, which is trained on randomly masked face photos with sampled sketches and color strokes in an adversarial manner. Xia et al.~\cite{xia2019cali} proposed a two-stage network for sketch-based portrait synthesis. The stroke calibration network is responsible for converting the input poorly-drawn sketch to a more detailed and calibrated one that resembles edge maps. Then the refined sketch is used in the image synthesis network to get a photo-realistic portrait image. Li et al.~\cite{li2019linestofacephoto} proposed a self-attention module to capture long-range connections of sketch structures, where self-attention mechanism is adopted to aggregate features from all positions of the feature map by the calculated self-attention map. A multi-scale discriminator is used to distinguish patches of different receptive fields, to simultaneously ensure local and global realism. Chen et al.~\cite{chen2020deepfacedrawing} introduced DeepFaceDrawing, a local-to-global approach for generating face images from sketches that uses input sketches as soft constraints and is able to produce high-quality face images even from rough and/or incomplete sketches.  The key idea is to learn feature embeddings of key face components and then train a deep neural network to map the embedded component features to realistic images.

While most works in sketch-to-image synthesis with deep learning techniques have focused on synthesizing object-level images from sketches, 
Gao et al.~\cite{gao_sketchycoco_2020} explored synthesis at the scene level by proposing a deep learning framework for scene-level image generation from freehand sketches. The framework first segments the sketch into individual objects, recognizes their classes, and categories them into foreground/background objects. Then the foreground objects are generated by an EdgeGAN module that learns a common vector representation for images and sketches and maps the vector representation of an input sketch to an image.  The background generation module is based on the pix2pix~\cite{isola2017image} architecture. The synthesized foregrounds along with background sketches are fed to a network to get the final generated scene. To train the network and evaluate their method, the authors constructed a composite dataset called SketchyCOCO based on the Sketchy database~\cite{sangkloy2016sketchy}, Tuberlin dataset~\cite{eitz2012humans}, QuickDraw dataset, and COCO Stuff~\cite{caesar2018cvpr}.

Considering that collecting paired training data can be labor intensive, learning from unpaired sketch-photo data in an unsupervised setting is an interesting direction to explore. 
Liu et al.~\cite{liu2020unsupervised} proposed an unsupervised solution by decomposing the synthesis process into a shape translation stage and a content enrichment stage.
The shape translation network transforms an input sketch into a gray-scale image, trained using unpaired sketches and images, under the supervision of a cycle-consistency loss. In the content enrichment stage, a reference image can be provided as style guidance, whose information is injected into the synthesis process following the AdaIN framework~\cite{huang2017arbitrary}.

\noindent {\bf Benchmark Datasets.} For synthesis from sketches, various datasets covering multiple types of objects are used~\cite{yu2016sketch,KrauseStarkDengFei-Fei_3DRR2013,finegrained,semjitter,WelinderEtal2010,liu2015faceattributes,karras2017progressive,sangkloy2016sketchy,wang2008face,lin2014microsoft,caesar2018cvpr}. However, only a few of them~\cite{yu2016sketch,sangkloy2016sketchy,wang2008face} have paired image-sketch data. For the other datasets, edge maps or line strokes are extracted using edge extraction or style transfer techniques and used as fake sketch data for training and validation. SketchyCOCO~\cite{gao_sketchycoco_2020} built a paired image-sketch dataset from existing image datasets~\cite{caesar2018cvpr} and sketch datasets~\cite{sangkloy2016sketchy,eitz2012humans} by looking for the most similar sketch with the same class label for each foreground object in a natural image. 
\vspace{5pt}
\subsubsection{Semantic Label Maps as Input}
\begin{figure}[ht]
    \centering
    \includegraphics[width=0.99\linewidth]{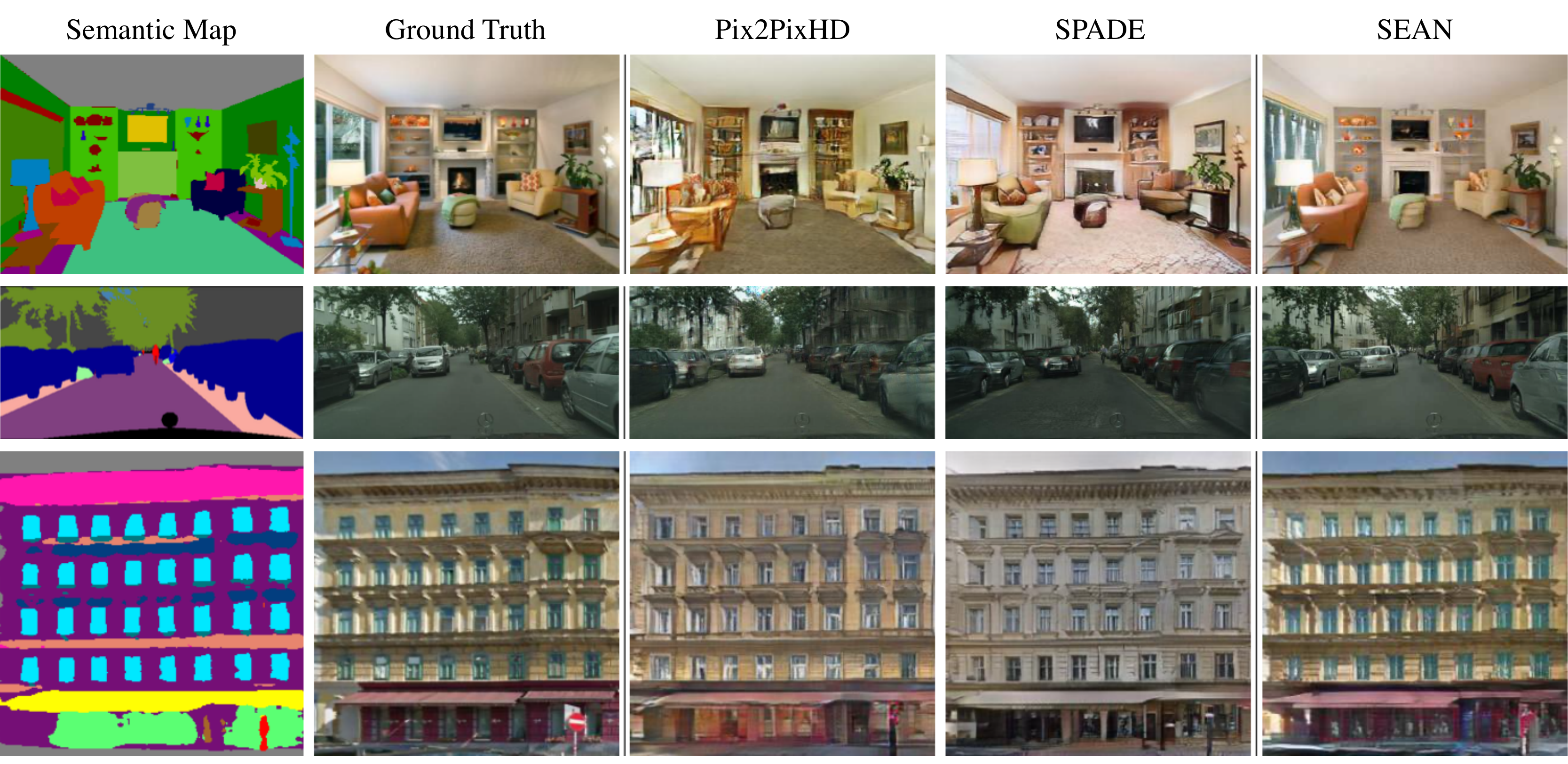}
    \caption{Illustration for image synthesis from semantic label maps. Image taken from~\cite{zhu2020sean}.}
    \label{fig:semantic}
\end{figure}
Synthesizing photorealistic images from semantic label maps is the inverse problem of semantic image segmentation. It has applications in controllable image synthesis and image editing. Existing methods either work with a traditional retrieval-and-composition approach
\cite{johnson_semantic_2006,bansal_shapes_2019}, a deep learning based method
\cite{chen_photographic_2017,lassner_generative_2017,park_semantic_2019,liu_learning_2020,zhu_semantically_2020,tang2020local}, or a hybrid of the two~\cite{qi_semi-parametric_2018}. Different types of datasets are utilized to allow synthesizing images of various scenes or subjects, such as indoor/outdoor scenes, or human bodies.

\noindent {\bf Retrieval-and-Composition based Methods.}
Non-parametric methods follow the traditional retrieval-and-composition strategy. 
Johnson et al.~\cite{johnson_semantic_2006} first proposed to synthesize images from semantic concepts. Given an empty canvas, the user can paint regions with corresponding keywords at desired locations. The algorithm searches for candidate images in the stock and uses a graph-cut based seam optimization process to generate realistic photographs for each combination. The best combination with the minimum seam cost is chosen as the final result.
Bansal et al.~\cite{bansal_shapes_2019} proposed a non-parametric matching and hierarchical composition strategy to synthesize realistic images from semantic maps.  The strategy consists of four stages: a global consistency stage to retrieve relevant samples based on indicator vectors of presented categories, a shape consistency stage to find candidate segments based on shape context similarity between the input label mask and the ones in the database, a part consistency stage and a pixel consistency stage that re-synthesize patches and pixels based on best-matching areas as measured by Hamming distance. 
The proposed method outperforms state-of-the-art parametric methods like pix2pix~\cite{isola2017image} and pix2pixHD~\cite{wang2018high} both qualitatively and quantitatively.

\noindent {\bf Deep Learning based Methods.}
Methods based on deep learning mainly vary in network architecture design and optimization objective.
Chen et al.~\cite{chen_photographic_2017} proposed a regression approach for synthesizing realistic images from semantic maps, without the need for adversarial training. To improve synthesis quality, they proposed a Cascaded Refinement Network (CRN), which progressively generates images from low resolution to high resolution (up to 2 megapixels at 1024x2048 pixel resolution) through a cascade of refinement modules. 
To encourage diversity in generated images, the authors proposed a diversity loss, which lets the network output multiple images at a time and optimize diversity within the collection. 
Wang et al.~\cite{wang2019example} proposed a style-consistent GAN framework that generates images given a semantic label map input and an exemplary image indicating style. A novel style-consistent discriminator is designed to determine whether a pair of images are consistent in style and an adaptive semantic consistency loss is optimized to ensure correspondence between the generated image and input semantic label map. 

Having found that directly synthesizing images from semantic maps through a sequence of convolutions sometimes provides non-satisfactory results because of semantic information loss during forward propagation, some works seek to better use the input semantic map and preserve semantic information in all stages of the synthesis network. 
Park et al.~\cite{park_semantic_2019} proposed a spatially-adaptive normalization layer (SPADE), which is a 
normalization layer with learnable parameters that utilizes the original semantic map to help retain semantic information in the feature maps after the traditional batch normalization. 
The authors incorporated their SPADE layers into the pix2pixHD architecture and produced state-of-the-art results on multiple datasets.
Liu et al.~\cite{liu_learning_2020} argues that the convolutional network should be sensitive to semantic layouts at different locations. Thus they proposed Conditional Convolution Blocks (CC Block), where parameters for convolution kernels are predicted from semantic layouts. They also proposed a feature pyramid semantics-embedding (FPSE) discriminator, which predicts semantic alignment scores in addition to real/fake scores. It explicitly forces the generated images to be better aligned semantically with the given semantic map.
Zhu et al.~\cite{zhu_semantically_2020} proposed a Group Decreasing Network (GroupDNet). GroupDNet utilizes group convolutions in the generator and the group number in the decoder decreases progressively. Inspired by SPADE, the authors also proposed a novel normalization layer to make better use of the information in the input semantic map. Experiments show that the GroupDNet architecture is more suitable for the multi-modal image synthesis (SMIS) task, and can produce plausible results.

Observing that results from existing methods often lack detailed local texture, resulting from large objects dominating the training, Tang et al.~\cite{tang2020local} aims for better synthesis of small objects in the image. In their design, each class has its own class-level generation network that is trained with feedback from a classification loss, and all the classes share an image-level global generator. The class-level generator generates parts of the image that correspond to each class, from masked feature maps. All the class-specific image parts are then combined and fused with the image-level generation result. 
In another work, to provide more fine-grained interactivity, Zhu et al.~\cite{zhu2020sean} proposed semantic region-adaptive normalization (SEAN), which allows manipulation of each semantic region individually, to improve image quality.

\noindent {\bf Integration methods. }
While deep learning based generative methods are better able to synthesize novel images, traditional retrieval-and-composition methods generate images with more reliable texture and less artifacts.
To combine the advantages of both parametric and non-parametric methods, Qi et al.~\cite{qi_semi-parametric_2018} presented a semi-parametric approach. They built a memory bank offline, containing segments of different classes of objects. Given an input semantic map, segments are first retrieved using a similarity metric defined by IoU score of the masks. The retrieved segments are fed to a spatial transformer network where they are aligned, and further put onto a canvas by an ordering network. The canvas is refined by a synthesis network to get the final result. This combination of retrieval-and-composition and deep-learning based methods allows high-fidelity image generation, but it takes more time during inference and the framework is not end-to-end trainable.

\noindent {\bf Benchmark Datasets.} For synthesis from semantic label maps, experiments are mainly conducted on datasets of human body~\cite{liang2015deep,liang2015human,liuLQWTcvpr16DeepFashion}, human face~\cite{CelebAMask-HQ}, indoor scenes~\cite{zhou2016semantic,zhou2017scene,Silberman:ECCV12} and outdoor scenes~\cite{cordts2016cityscapes}. Lassner \textit{et al.}~\cite{lassner_generative_2017} augmented the Chictopia10K~\cite{liang2015deep,liang2015human} dataset by adding 2D keypoint locations and fitted SMPL body models, and the augmented dataset is used by Bem \textit{et al.}~\cite{de2019conditional}. Park \textit{et al.}~\cite{park_semantic_2019} and Zhu \textit{et al.}~\cite{zhu2020sean} collected images from the Internet and applied state-of-the-art semantic segmentation models~\cite{chen2017deeplab,chen2018encoder} to build paired datasets.


\vspace{5pt}
\subsubsection{Poses as Input}

Given a reference person image, its corresponding pose, and a novel pose, pose-based image synthesis methods can generate an image of the person in that novel pose. Different from synthesizing images from sketches or semantic maps, pose-guided synthesis requires novel views to be generated, which cannot be done by the retrieval and composition pipeline.  Thus we focus on reviewing deep learning-based methods~\cite{balakrishnan_synthesizing_2018,ma_pose_2018,ma_disentangled_2018,siarohin_deformable_2018,pumarola2018unsupervised,de2019conditional,dong2018soft,li2019dense,song2019unsupervised,zhu2019progressive}.  In these methods, a pose is often represented as a set of well-defined body keypoints. Each of the keypoints can be modeled as an isotropic Gaussian that is centered at the ground-truth joint location and has a small standard deviation, giving rise to a heatmap.  The concatenation of the joint-centered heatmaps then can be used as the input to the image synthesis network. Heatmaps of rigid parts and the whole body can also be utilized~\cite{de2019conditional}. 

\noindent {\bf Supervised Deep Learning Methods.}
In a supervised setting, ground truth target images under target poses are required for training. Thus, datasets with the same person in multiple poses are needed.
Ma et al.~\cite{ma_pose_2018} proposed the Pose Guided Person Generation Network for generating person images under given poses. It adopts a GAN-like architecture and generates images in a coarse-to-fine manner.
In the coarse stage, an image of a person along with a novel pose are fed into the U-Net based generator, where the pose is represented as heatmaps of body keypoints. The coarse output is then concatenated again with the person image, and a refinement network is trained to learn a difference map that can be added to the coarse output to get the final refined result. The discriminator is trained to distinguish  synthesized outputs and real images. Besides the GAN loss, an L1 loss is used to measure dissimilarity between the generated output and the target image. Since the target image may have different background from the input condition image, the L1 loss is modified to give higher weight to the human body utilizing a pose mask derived from the pose skeleton. 

Although GANs have achieved great success in image synthesis, there are still some difficulties when it comes to pose-based synthesis, one of which being the deformation problem. The given novel pose can be drastically different from the original pose, resulting in large deformations in both shape and texture in the synthesized image and making it hard to directly train a network that is able to generate images without artifacts.
Existing works mainly adopt transformation strategies to overcome this problem, because transformation makes it explicit about which body part will be moved to which place, 
being aware of the original and target poses.   These methods usually transform body parts of the original image~\cite{balakrishnan_synthesizing_2018}, the human parsing map~\cite{dong2018soft}, or the feature map~\cite{siarohin_deformable_2018,dong2018soft,zhu2019progressive}.
Balakrishnan et al.~\cite{balakrishnan_synthesizing_2018} explicitly separate the human body from the background and synthesize person images of unseen poses and background in separate steps. 
Their method consists of four modules: a segmentation module that produces masks of the whole body and each body part based on the source image and pose; a transformation module that calculates and applies affine transformation to each body part and corresponding feature maps; a background generation module that applies inpainting to fill the body-removed foreground region; and a final integration module that uses the transformed feature maps and the target pose to get the synthesized foreground, which is then combined with the inpainted background to get the final result. 
To train the network, they use a VGG-19 perceptual loss along with a GAN loss.
Siarohin et al.~\cite{siarohin_deformable_2018} noted that it is hard for the generator to directly capture large body movements because of the restricted receptive field, and introduced deformable GANs to tackle the problem. 
The method decomposes the body joints into several semantic parts, and calculates an affine transform from the source to the target pose for each part. The affine transforms are used to align the feature maps of the source image with the target pose. The transformed feature maps are then concatenated with the target pose features and decoded to synthesize the output image. The authors also proposed a novel nearest-neighbor loss based on feature maps, instead of using L1 or L2 loss. Their method is more robust to large pose changes and produces higher quality images compared to~\cite{ma_pose_2018}.
Dong et al.~\cite{dong2018soft} utilize parsing results as a proxy to achieve better synthesizing results. They first estimate parsing results for the target pose, then fit a Thin Plate Spline (TPS) transformation between the original and estimated parsing maps. The TPS transformation is further applied to warp the feature maps for feature alignment and a soft-gated warping block is developed to provide controllability to the transformation degree. The final image is synthesized based on the transformed feature maps.
Zhu et al.~\cite{zhu2019progressive} proposed that large deformations can be divided into a sequence of small deformations, which are more friendly to network training. In this way, the original pose can be transformed progressively, through many intermediate poses. They proposed a Pose-Attentional Transfer Block (PATB), which transforms the feature maps under the guidance of an attention mask. By stacking multiple PATBs, the feature maps undergo several transformations and the transformed maps are used to synthesize the final result.

While most of the deep learning based methods for synthesis from poses adopt an adversarial training paradigm, Bem et al.~\cite{de2019conditional} 
proposed a conditional-VAEGAN architecture that combines a conditional-VAE framework and a GAN discriminator module to generate realistic natural images of people in a unified probabilistic framework where the body pose and appearance are kept as separated and interpretable variables, allowing the sampling of people with independent variations of pose and appearance. The loss function used includes both conditional-VAE and GAN losses composed of L1 reconstruction loss, closed-form KL-divergence loss between recognition and prior distributions, and discriminator cross-entropy loss. 


\noindent {\bf Unsupervised Deep Learning Methods.}
The aforementioned pose-to-image synthesis methods require ground truth images under target poses for training because of their use of L1, L2 or perceptual losses.
To eliminate the need for target images, 
some works focus on the unsupervised setting of this problem~\cite{pumarola2018unsupervised,song2019unsupervised}, where the training process does not require ground truth image of the target pose.
The basic idea is to ensure cycle consistency. After the forward pass, the synthesized result along with the target pose will be treated as the reference, and be used to synthesize the image under the original reference pose. This synthesized image should be consistent with the original reference image. Pumarola et al.~\cite{pumarola2018unsupervised} further utilize a pose estimator, to ensure pose consistency. Song et al.~\cite{song2019unsupervised} use parsing maps as supervision instead of poses. They predict parsing maps under new target poses and use them to synthesize the corresponding images. Since the parsing maps under the target poses are not available due to operating in the unsupervised setting, the authors proposed a pseudo-label selection technique to get ``fake" parsing maps by searching for the ones with the same clothes type and minimum transformation energy. 

\noindent {\bf Benchmark Datasets.} For synthesis from poses, the DeepFashion~\cite{liuLQWTcvpr16DeepFashion} and Market-1501~\cite{zheng2015scalable} datasets are most widely used.  The DeepFashion dataset is built for clothes recognition but has also been used for pose-based image synthesis because of the rich annotations available such as clothing landmarks as well as images with corresponding foreground but diverse backgrounds. The Market-1501 dataset was initially introduced for the purpose of person re-identification, and it contains a large number of person images produced using a pedestrian detector and annotated bounding boxes; also, each identity has multiple images from different camera views.

\subsection{Other Input Modalities}
Except for text descriptions and image-like inputs, there are other intuitive user inputs such as class labels, attribute vectors, and graph-like inputs. 

\subsubsection{Visual Attributes as Input}\label{app:attribute}

In this subsection, we mainly focus on works that use one of the fine-grained class conditional labels or vectors, \textit{i.e.} visual attributes, as inputs. Visual attributes provide a simple and accurate way of describing major features present in images, such as in describing attributes of a certain category of birds or details of a person's  face. Current methods either take a discrete one-hot vector as attribute labels, or a continuous vector as visual attribute input. 

Yan~\textit{et al.}~\cite{yan2016attribute2image} proposes a disentangling CVAE (disCVAE) for  conditioned image generation from visual
attributes. While conditional Variational Auto-Encoder (cVAE)~\cite{sohn2015learning} generates images from the posterior conditioned on both the conditions and random vectors, disCVAE interprets an image as a composite of a foreground layer and a background layer. The foreground layer is conditioned on visual attributes and the whole image is generated through a gated integration.  Attribute-conditioned experiments are often conducted on the LFW~\cite{LFWTech} and CUB~\cite{WelinderEtal2010} datasets.

For face generation with visual attribute inputs, one related application is manipulating existing face images with provided attributes. AttGAN~\cite{he2019attgan} applies attribute classification constraint and reconstruction learning to guarantee the change of desired attributes while maintaining other details. Zhang~\textit{et al.}~\cite{zhang2018generative} proposes spatial attention which can localize attribute-specific regions to perform desired attribute manipulation and keep the rest unchanged. Unlike other works utilizing attributes input, Qian~\textit{et al.}~\cite{qian2019make} explores face manipulation via conditional structure input. Given structure prior as conditional input of the cVAE, AF-VAE~\cite{qian2019make} can arbitrarily modify facial expressions and head
poses using geometry-guided feature disentanglement and additive Gaussian Mixture prior for appearance representation. Most such face image manipulation works perform experiments on commonly used face image datasets such as the CelebA~\cite{liu2015faceattributes} dataset.

For controllable person image synthesis, Men~\textit{et al.}~\cite{men2020controllable} introduces Attribute-Decomposed GAN, where visual attributes including clothes are extracted from reference images and combined with target poses to generate target images with desired attributes. The separation and decomposition of attributes from existing images provide a new way of synthesizing person images without attribute annotations.

Another interesting application of taking visual attributes as input is fashion design. Lee~\textit{et al.}~\cite{lee2019fashion} proposes a GAN model with an attentional
discriminator for attribute-to-fashion generation. For multiple-attribute inputs, multiple independent Gaussian
distributions are derived by mapping each attribute vector
to the mean vector and diagonal covariance matrix. The prior distribution for attribute combination is the product of all independent Gaussians. Experiments are conducted on a dataset consisting of dress images collected from a popular fashion site.

In terms of image generation methodology using visual attributes as inputs, the Glow model introduced in ~\cite{kingma2018glow} as a generative flow model using an invertible $1\times1$ convolution shows great potentials. Compared with VAEs and GANs, flow models have merits including  reversible generation, meaningful latent space, and memory efficiency. Glow consists of a series of steps of flow, where each step consists of activation normalization followed by an invertible $1\times1$ convolution, then followed by a coupling layer. On the Cifar10 dataset, Glow achieves better negative log likelihood than RealNVP~\cite{dinh2016density}. On the CelebA-HQ dataset, Glow generates high fidelity face images and also allows meaningful visual attribute manipulation.

\noindent {\bf Benchmark Datasets.} For attributes-guided synthesis tasks, major benckmarking datasets include Visual Genome, CelebA(-HQ), and Labeled Faces in the Wild. Visual Genome~\cite{krishna2017visual} contains over 100K images where each image has an average of 21 objects, 18 attributes, and 18 pairwise relationships between objects. The CelebA~\cite{liu2015faceattributes} dataset has a 40 dimensional binary
attribute vector annotated for each face image. The CelebA-HQ dataset~\cite{karras2017progressive} consists of 30,000 high resolution images from the CelebA dataset. The Labeled Faces in
the Wild (LFW) dataset contains face images that are segmented and labeled with semantically meaningful region labels (e.g., hair, skin).

\subsubsection{Graphs and Layouts as Input}
\begin{figure}[tbp]
    \centering
    \includegraphics[width=0.95\linewidth]{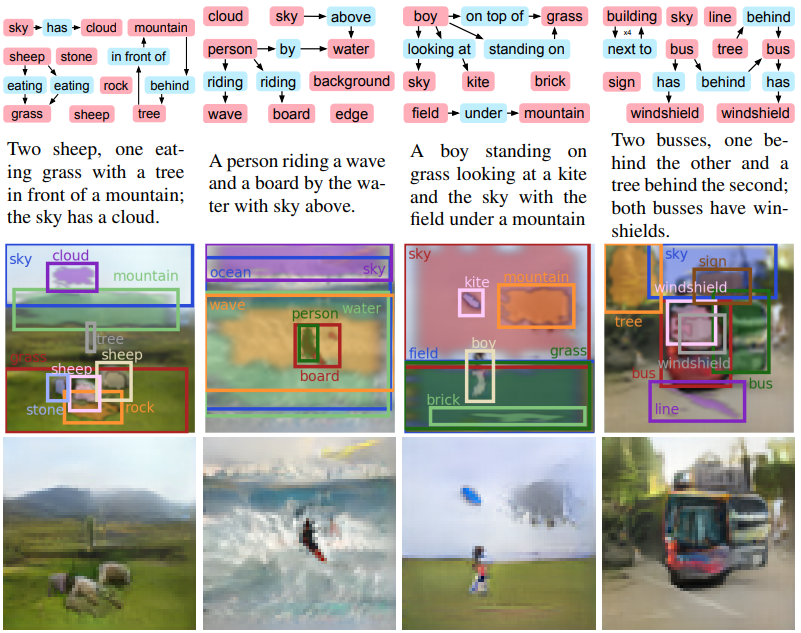}
    \caption{Example scene graph to image synthesis results. Scene graphs are often extracted from text descriptions. Correct object relationships embedded in input scene graphs are reflected in the generated images. Image taken from ~\cite{johnson2018image}.}
    \label{fig:scene_graph}
\end{figure}

Another interesting type of intuitive user input is graphs (Fig.~\ref{fig:scene_graph}). Graphs can encode multiple relationships in a concise way and have very unique characteristics such as sparse representation. An example application of graph-based inputs is architecture design using scene graphs, layouts, and other similar modalities.

Johnson~\textit{et al.}~\cite{johnson2018image}, as mentioned earlier in Section \ref{sec:text-layout}, can take a scene graph and generate the corresponding layout. The final image is then synthesized by a CRN model~\cite{chen2017photographic} from a noise vector and the layout. Figure \ref{fig:scene_graph} demonstrates some results from \cite{johnson2018image}.

To generate images that exhibit complex relationships among multiple objects, Zhao~\textit{et al.}~\cite{zhao2019image} proposes a Layout2Im model that uses layout as input to generate images. The layout is specified by multiple bounding boxes of objects with category labels. 
Training of the model is done by taking groundtruth images with their layouts, and testing is done by sampling object latent codes from a normal distribution. An object composer takes the word embedding of input text, object latent code, and bounding box locations to composite object feature maps. The object feature maps are then composed using convolutional LSTM into a hidden feature map and decoded into the final image.

Also containing the idea of converting layout to image, LayoutGAN~\cite{li2019layoutgan} uses a differentiable wireframe rendering layer with an image-based discriminator that can generate layout from graphical element inputs. Semantic and spatial relations between elements are learned via a stacked relation module with self attention, and experiments on various datasets show promising results in generating meaningful layouts which can be also rasterized. 

Luo~\textit{et al.}~\cite{Luo_2020_CVPR} proposes a variational generative model which generates 3D scene layouts given input scene graphs. cVAE is combined with the graph convolution network (GCN)~\cite{kipf2016semi} for layout synthesis. The authors also present a rendering model which first instantiates a 3D model by retrieving object meshes, then utilizes a differentiable renderer to render the corresponding
semantic image and the depth image. Their experiments on the SUNCG dataset~\cite{song2017semantic} show that the method can generate accurate and diverse 3D scene layouts and has potential in various downstream scene layout and image synthesis tasks.


\section{Summary and Trends}\label{sec:evaluation}
\subsection{Advances in Model Architecture Design and Training Strategy}
Among different attempts of improving the synthesized image quality and the correspondence between user input and generated image, several successful designs are incorporated into multiple conditional generative models and have proven their effectiveness in various tasks. For instance, a hierarchical generation architecture has been widely used by different models, including GANs~\cite{zhang2018stackgan++, choi2018stargan, wang2018high} and VAEs~\cite{vahdat2020NVAE}, in order to generate high-resolution, high-quality images in a multi-stage, progressive fashion. Attention-based mechanisms are proposed and incorporated in multiple works \cite{xu2018attngan, zhang2019self} towards more fine-grained control over regions within generated images. To ensure correspondence between user input and generated images, various designs are proposed for generative neural networks: Relatively straightforward methods take the combination of user input and other input (e.g., latent vector) as input to the generative model; other methods take the user input as part of the supervision signal to measure the correspondence between input and output; more advanced methods, which may also be more effective, combine transformed inputs together, such as in projection discriminator~\cite{miyato2018cgans} and spatially-adaptive normalization ~\cite{park_semantic_2019}.

While most of the current successful models are based on GANs, it is well-known that GAN training is difficult and can be unstable. Similar to general purpose GANs, works focusing on image synthesis with intuitive user inputs also adopt different design and training strategies to ease and stabilize the GAN training. Commonly used normalisations include conditional batch normalization~\cite{de2017modulating} and spectral normalization~\cite{miyato2018spectral}; commonly used adversarial losses include WGAN loss with different regularizations~\cite{arjovsky2017wasserstein, gulrajani2017improved}, LS-GAN loss~\cite{mao2017least} and Hinge loss~\cite{lim2017geometric}. To balance the training of the generator and the discriminator, imbalanced training strategies such as two time-scale update rule (TTUR)~\cite{heusel2017gans} have also been adopted for better convergence. 

General losses employed in different models heavily depend on the methodological framework. Retrieval and composition methods typically do not need to be trained, therefore no loss is used. For GAN-like models, an adversarial loss is essential in a majority of the models, which combines a loss for the generator and a loss for the discriminator in order to push the generator toward generating fake samples that match the distribution of real examples. Widely used adversarial losses include the minimax loss introduced in the original GAN paper \cite{goodfellow2014generative} and the Wasserstein loss introduced in the WGAN paper \cite{arjovsky2017wasserstein}.
VAE models are typically trained by minimizing a reconstruction error between the encoder-decoded data and the initial data, with some regularization of the latent space \cite{kingma2013auto}. To evaluate the visual quality of generated images and optimize toward better image quality, 
perceptual loss \cite{johnson2016perceptual} or adversarial feature matching loss \cite{salimans2016improved} have been adopted by many existing works, especially when paired supervision signal is available. 

Alongside the general losses, auxiliary losses are often incorporated in models to better handle different tasks. Task-specific losses, as well as evaluation metrics, are natural choices to evaluate and improve task-specific performances. Depending on the output modalities, one commonly used loss or metric is to recover the input condition from the synthesized images. For instance, image captioning losses can be included in text-to-image synthesis models \cite{qiao2019mirrorgan}, and pose prediction losses can complement the general losses in pose-to-image synthesis tasks \cite{pumarola2018unsupervised,song2019unsupervised}. 

\subsection{Summary on Methods using Specific Input Types}
Recent advances in text-to-image synthesis have been mainly based on deep learning methods, especially GANs. Two major challenges of the text-to-image synthesis task are learning the correspondence between text descriptions and generated images, and ensuring the quality of generated images. 
The text-image correspondence problem has been addressed in recent years with advanced embedding techniques of text descriptions and special designs such as attention mechanisms used to match words and image regions.
For the quality of generated images, however, promising results are still limited to generating narrow categories of objects. For general scenes where multiple objects co-exist with complex relationships, the realism and diversity of the generated images are not satisfactory and remain to be improved. To reduce the difficulty of synthesizing complex scenes, current models may benefit from leveraging different methods such as combining retrieval-and-composition with deep learning, and relationship learning which uses relation graphs as auxiliary input or intermediate step. 

For image-like inputs, one can take a traditional retrieval-and-composition strategy or adopt the more recent deep learning based methods.
The retrieval-and-composition strategy has several advantages. First, its outputs contain fewer artifacts because the objects are retrieved rather than synthesized. Second, it is more user-friendly, since it allows user intervention in all stages of the workflow, which brings controllability and customizability. Third, it can be directly applied to a new dataset, without the need for time-consuming training or adaptation. In comparison, deep learning based methods are less interpretable and more difficult to accept user intervention in all stages of the synthesis process. Although some attempts in combining the advantages of the two approaches have been made~\cite{qi_semi-parametric_2018}, deep-learning based methods still dominate for their versatility and ability to generate completely novel images. 
In these deep learning based methods, inputs are usually represented as regular grid structures like rasterized images (e.g. for sketches) or multi-channel tensors (e.g. for poses, semantic maps), for the convenience of utilizing convolution based neural networks. 
Methods for different input types also have their own emphases. Works for sketch-based synthesis have attempted to bridge the gap between synthesized sketches and real free-hand sketches, because the latter is hard to collect and synthesized sketches can be used to satisfy the needs of training large networks. For synthesis based on semantic maps, progress has been made mainly on the design of network architectures in order to better utilize information in the input semantic maps.
For pose-based synthesis, various solutions are proposed to address problems caused by large deformations between source and target poses, including 
performing explicit transformations, learning pixel-level correspondence, and synthesizing through a sequence of mild deformations. 
Efforts have also been made to alleviate the need for ground-truth data in supervised learning settings.
Take pose-based synthesis for example, the supervised setting requires multiple images of the same person with the same background but different poses; however, what we often have is an image collection with only one image for each person. Some methods \cite{pumarola2018unsupervised,song2019unsupervised,liu2020unsupervised} are proposed to work under an unsupervised setting, where no ground-truth of the synthesized result is needed; they mainly work by constraining cycle consistency, with extra supervision for intermediate outputs.

For image synthesis with visual attributes, applications in the reviewed works have been mainly on face synthesis, person synthesis, and fashion design. Since attributes are an intuitive type of user input suitable for interactive synthesis, we believe that more applications should be  explored and more advanced models can be proposed. One bottleneck for current visual attribute based synthesis tasks is that attribute-level annotations are often required for supervised training. For datasets with no attribute-level annotations, unsupervised attribute disentanglement or attribute-related prior knowledge need to be incorporated into the model design to guarantee that the generated images have the correct attributes.

Image synthesis with graphs as input can better encode relationships between objects than using other intuitive user inputs. Current works often rely on graph neural networks~\cite{kipf2016semi, velivckovic2018graph} to learn the graph and node features. In addition to using graphs as input, current methods also try to generate scene graphs as intermediate output from other modalities of input such as text descriptions. Applications of using graphs as intuitive input include architecture design and scene synthesis that require the preservation of specific object relationships. While fewer works have been done for image synthesis with graphs, we believe it has great potential in advancing techniques capable of generating scenes with multiple objects, complex relationships, and structural constraints. 





\begin{table*}[ht]
\begin{minipage}{\linewidth}
\resizebox{\linewidth}{!}{
    \begin{tabular}{ |c||c|c|c|c|c| }
    \hline
    \textbf{Dataset name} & \textbf{\# images} & \textbf{Categories} & \textbf{Annotations} & \textbf{Tasks} & \textbf{Used in}\\\hline
     Shoe V2~\cite{yu2016sketch} & 8,648\footnote{2,000 real images and 6,648 sketches.} & shoe & P & SK & \cite{liu2020unsupervised} \\\hline
     Stanford's Cars~\cite{KrauseStarkDengFei-Fei_3DRR2013} & 16,185 & car & L,BB & SK & \cite{lu_image_2018} \\\hline
     UT Zappos50K~\cite{finegrained,semjitter} & 50,025 & shoe & L,P & SK & \cite{ghosh_interactive_2019} \\\hline
     Caltech-UCSD Birds 200~\cite{WelinderEtal2010} & 6,033 & bird & L,A,BB,S & TE, SK & \cite{yan2016attribute2image, reed2016generative, zhang2017stackgan, zhang2018stackgan++, xu2018attngan, zhang2018photographic, bodla2018semi, yin2019semantics, zhu2019dm, qiao2019mirrorgan, lu_image_2018} \\\hline
     Oxford-102~\cite{nilsback2008automated} &  8,189 & flower & L & TE & \cite{reed2016generative, zhang2017stackgan, zhang2018stackgan++, xu2018attngan, zhang2018photographic, zhu2019dm, qiao2019mirrorgan} \\\hline
     Labeled Faces in the Wild~\cite{LFWTech} & 13,233 & face & L,S & AT & \cite{yan2016attribute2image,zhang2018generative}\\\hline
     CelebA~\cite{liu2015faceattributes} & 202,599 & face & L,A,KP & SK, AT & \cite{zhang2018generative,qian2019make,he2019attgan,lu_image_2018, bodla2018semi} \\\hline
     CelebA-HQ~\cite{karras2017progressive} & 30,000 & face & L,A,KP & SK, AT & \cite{kingma2018glow,portenier2018faceshop,li2019linestofacephoto} \\\hline
     Sketchy~\cite{sangkloy2016sketchy} & 87,971\footnote{12,500 real images and 75,471 sketches.} & objects & L,P & SK & \cite{chen_sketchygan_2018} \\\hline
     CUHK Face Sketch~\cite{wang2008face} & 1212\footnote{606 pairs of real face photo and the corresponding sketch.} & face & P & SK & \cite{gucluturk_convolutional_2016,sangkloy_scribbler_2017,xia2019cali} \\\hline
     COCO~\cite{lin2014microsoft} & 330,000 & objects & BB,S,KP,T & TE,SK,SE & \cite{mansimov2015generating, zhang2017stackgan, zhang2018stackgan++, xu2018attngan, zhang2018photographic, hong2018inferring, li2019object, tan2018text2scene, yin2019semantics, zhu2019dm, qiao2019mirrorgan, hinz2019semantic, wang_mindcamera_2018,bansal_shapes_2019} \\\hline
     COCO-Stuff~\cite{caesar2018cvpr} & 164,000 & objects & S,C & SK,SE,SG,LA & \cite{zhao2019image,johnson2018image,gao_sketchycoco_2020,park_semantic_2019,liu_learning_2020} \\\hline
     CelebAMask-HQ~\cite{CelebAMask-HQ} & 30,000 & face & S & SE & \cite{zhu2020sean} \\\hline
     Cityscapes~\cite{cordts2016cityscapes} & 25,000 & outdoor scene & S & SE & \cite{chen_photographic_2017,qi_semi-parametric_2018,park_semantic_2019,liu_learning_2020,zhu_semantically_2020,tang2020local,zhu2020sean}\\\hline
     ADE20K~\cite{zhou2016semantic,zhou2017scene} & 22,210 & indoor scene & S & SE & \cite{qi_semi-parametric_2018,park_semantic_2019,liu_learning_2020,zhu_semantically_2020,tang2020local,zhu2020sean} \\\hline
     NYU Depth~\cite{Silberman:ECCV12} & 1,449 & indoor scene & S,D & SE & \cite{chen_photographic_2017,qi_semi-parametric_2018} \\\hline
     Chictopia10K~\cite{liang2015deep,liang2015human} & 17,706 & human & S & SE & \cite{lassner_generative_2017} \\\hline
     DeepFashion~\cite{liuLQWTcvpr16DeepFashion} & 52,712 & human & L,A,P,KP & SE,P,AT & \cite{zhu_semantically_2020,ma_pose_2018,ma_disentangled_2018,siarohin_deformable_2018,pumarola2018unsupervised,dong2018soft,li2019dense,song2019unsupervised,zhu2019progressive,men2020controllable} \\\hline
     Market-1501~\cite{zheng2015scalable} & 32,668 & human & L,A & P & \cite{ma_pose_2018,ma_disentangled_2018,siarohin_deformable_2018,dong2018soft,li2019dense,song2019unsupervised,zhu2019progressive} \\\hline
     Human3.6M~\cite{ionescu2013human3} & 3,600,000 & human & KP,BB,S,SC & P & \cite{de2019conditional} \\\hline
     Visual Genome~\cite{krishna2017visual} & 108,077 & objects & BB,A,R,T,VQA & SG,LA & \cite{zhao2019image,johnson2018image} \\\hline
    \end{tabular}
}
\vspace{0.2cm}
\end{minipage}
\caption{Commonly used datasets in image synthesis tasks with intuitive user inputs. For annotations, possible values are \textbf{L}abel, \textbf{A}ttribute, \textbf{P}air, \textbf{K}ey\textbf{P}oint, \textbf{B}ounding \textbf{B}ox, \textbf{S}emantic map, \textbf{R}elationship, \textbf{T}ext, \textbf{V}isual \textbf{Q}uestion \textbf{A}nswers, \textbf{D}epth map, 3D \textbf{SC}an. For tasks, possible values are \textbf{TE}xt, \textbf{P}ose, \textbf{SK}etch, \textbf{SE}mantic map, \textbf{AT}tributes, \textbf{S}cene \textbf{G}raph, \textbf{LA}yout.}
\label{tab:datasets}
\end{table*}

\subsection{Summary on Benchmark Datasets}\label{sec:datasets}
To facilitate the lookup of datasets available for particular tasks or particular types of input, we summarize popular datasets used for various image synthesis tasks with intuitive user inputs in Table~\ref{tab:datasets}. 
State-of-the-art image synthesis methods have achieved high-quality results using datasets containing single object categories such as cars~\cite{KrauseStarkDengFei-Fei_3DRR2013}, birds~\cite{WelinderEtal2010}, and human faces~\cite{liu2015faceattributes,karras2017progressive,wang2008face,CelebAMask-HQ}. For synthesizing images that contain multiple object categories and complex scene structures, there is still room for improvement using datasets such as the MS-COCO~\cite{lin2014microsoft}. Future work can also focus more on synthesis with intuitive and interactive user inputs, as well as applications of the synthesis methods in real-world scenarios.

\section{New Perspectives}\label{sec:new-perspectives}
Having reviewed recent works for image synthesis given intuitive inputs, we discuss in this section new perspectives on future research that relate to input versatility, generation methodology, benchmark datasets and evaluation metrics.

\subsection{Input Versatility}

\noindent {\bf Text to Image.} While current methods for text-to-image synthesis mainly take text inputs that describe the visual content of an image, more natural inputs often contain affective words such as happy or pleasing, scary or frightful. To handle such inputs, it is necessary for models to consider the emotional effects as part of the input text comprehension. Further, generating images that express or incur a certain sentiment will require learning the mapping between visual content and emotional dimensions such as valence (i.e. positive or negative affectivity) and arousal (i.g. how calming or exciting the information is), as well as understanding how different compositions of the same objects in an image can lead to different sentiments.

For particular application domains, input text descriptions may be more versatile. For instance, in medical image synthesis, a given input can be a clinical report that contains one or several paragraphs of text description.  Such domain-specific inputs also require prior knowledge for input text comprehension and text-to-image mapping.   Other under-explored applications include taking paragraphs or multiple sentences as input to generate a sequence of images for story telling~\cite{li2019storygan}, or text-based video synthesis and editing~\cite{pan2017create, li2018video,wang2019write}.

For conditional synthesis, most current works perform one-to-many generation and try to improve the diversity of images generated given the same text input. One interesting work for text-to-image synthesis by Yin~\textit{et al.} proposes SD-GAN~\cite{yin2019semantics} which investigates the variability among different inputs intended for the same target image. New applications may be discovered that need methods for many-to-one synthesis using similar pipelines.

\noindent {\bf Image from sketch, pose, graphic inputs, and others.} 
For sketches and poses as user inputs, existing methods treat them as rasterized images to perform an image-to-image translation as the synthesis method. Considering that sketches and poses all contain geometry information and the relationships among different points on the geometry are important, we believe it is beneficial to investigate representing such inputs as sparse vectorized representations such as graphs, instead of using rasterized representations. 
Taking vectorized inputs will greatly reduce the input sizes and will also enable the use of existing graph understanding techniques such as graph neural networks. For sketches as input, another interesting task is to generate videos from sketch-based storyboards, since it has numerous applications in animation and visualization. 

For graphic inputs that represent architectural structures such as layouts and wireframes, an important consideration is that the synthesized images should preserve structural constraints such as junctions, parallel lines, and planar surfaces~\cite{xue2020neural} or relations between graphical elements~\cite{li2019layoutgan}. In these scenarios, incorporating prior knowledge about the physical world can help enhance the photorealism of generated images and improve the structural coherence of generated designs.

 It will also be interesting to further investigate image and/or video generation from other forms of inputs. Audio, for instance, is another intuitive, interactive and expressive type of input.   Generating photo-realistic video portraits that are in synch with input audio streams ~\cite{chen2019hierarchical,zhou2019talking,wen2020photorealistic} has many applications such as assisting the hearing impaired with speech comprehension, privacy-preserving video chat, and VR/AR for training professionals.

\subsection{Connections and Integration between Generation Paradigms}
In conditional image synthesis, deep learning based methods have been dominating and have shown promising results. However, they still have limitations including the requirement of large training datasets and high computational cost for training. 
Since the retrieval-and-composition methods are often light-weight and require little training, they can be complementary to the deep learning based methods. Existing works on image synthesis from semantic maps have explored the strategy of combining retrieval-and-composition and learning-based models~\cite{qi_semi-parametric_2018}. One way of combination could be using retrieval-and-composition to generate a draft image and then refining 
the image for better visual quality and diversity using a learning-based approach. 

Besides the quality of generated images, the controllability of the output and the interpretability of the model also play essential roles in the synthesis process. Although GAN models generally achieve better image quality than other methods, it is often more difficult to perform interactive or controllable generation using GAN methods than other learning based methods. Hybrid models such as the combination of GANs and VAEs~\cite{larsen2016autoencoding, mescheder2017adversarial, bao2017cvae,de2019conditional} have shown promising synthesis results as well as better feature disentanglement properties. Future works in image synthesis given intuitive user input can explore more possibilities of using hybrid models combining the advantages of GANs and VAEs such as in \cite{de2019conditional} as well as using normalizing flow based methods \cite{rezende2015variational,kingma2018glow} that allow both feature learning and tractable marginal likelihood estimation.

Overall, we believe cross pollination between major image generation paradigms will continue to be an important direction, which can produce new models that improve upon existing image synthesis paradigms by combining their merits and overcoming their limitations.




\subsection{Evaluation and comparison of generation methods}

\noindent {\bf Evaluation Metrics.} 
While a range of quantitative metrics for measuring the realism and diversity of generated images have been proposed including widely used IS~\cite{salimans2016improved}, FID~\cite{heusel2017gans}, and SSIM~\cite{wang2004image}, they are still lacking in consistency with human perception and that is why many works still rely on qualitative human evaluation to assess the quality of images synthesized by different methods. Recently, some metrics, such as R-precision~\cite{xu2018attngan} and SOA score~\cite{hinz2019semantic} in text-to-image synthesis, have been proposed to evaluate whether a generated image is well conditioned on the given input and try to achieve better consistency with human perception.  Further work on automatic metrics that match well with human evaluation will continue to be important.

For a specific task or application, evaluation should be based on not just the final image quality but how well the generated images match the conditional input and serve the purpose of the intended application or task.  If the synthesized images are used in down-stream tasks such as data augmentation for classification, evaluation based on down-stream tasks also provides valuable information. 

While it is difficult to compare methods across input types due to differences in input modality and interactivity, it is feasible to establish standard processes for synthesis from a particular kind of input, thus making it possible for fair comparison between methods given the same type of input using the same benchmark. 

\noindent {\bf Datasets.}
As shown in Sec. \ref{sec:datasets}, large-scale datasets of natural images and annotations have been collected for specific object categories such as human bodies, faces, birds, cars, and for scenes that contain multiple object categories such as those in COCO~\cite{lin2014microsoft} and CityScapes~\cite{cordts2016cityscapes}. As future work, in order to enable applications in particular domains that benefit from image synthesis such as medical image synthesis for data augmentation and movie video generation, domain-specific datasets with appropriate annotations will need to be created.   

\noindent {\bf Evaluation of input choices.}
Existing image generation methods have been evaluated and compared mainly based on their output, i.e. the generated images.  We believe that in image generation tasks conditioned on intuitive inputs, it is equally important to compare methods based on their input choice.  In Sec.~\ref{sec:input-characteristics}, we introduced several characteristics that can be used to compare and evaluate inputs such as their accessibility, expressiveness, and interactivity. It will be interesting to study other important characteristics of inputs as well as criteria for evaluating how well an input type meets the needs of an application, how well the input supports interactive editing, how regularized the learned latent space is, and how well the synthesized image matches the input condition. 

\section{Conclusions}\label{sec:conclusion}

This review has covered main approaches for image synthesis and rendering given intuitive user inputs. First, we examine what makes a good paradigm for image synthesis from intuitive user input, from the perspective of user input characteristics and that of output image quality.  We then provide an overview of main generation paradigms: retrieval and composition, cGAN, cVAE, and hybrid models, autoregressive models, normalizing flow based methods. Their relative strengths and weaknesses are discussed in hope of inspiring ideas that draw connections between the main approaches to produce models and methods that take advantage of the relative strengths of each paradigm. After the overview, we delve into details of specific algorithms for different input types and examine their ideas and contributions. In particular, we conduct a comprehensive literature review on approaches for generating images from text, sketches or strokes, semantic label maps, poses, visual attributes, graphs and layouts. Then, we summarize these existing methods in terms of benchmark datasets used and identify trends related to advances in model architecture design and training strategy, and strategies for handling specific input types. Last but not least, we provide our perspective on future directions related to input versatility, generation methodology, benchmark datasets, and method evaluation and comparison.


\bibliographystyle{plain}
\bibliography{ref}

\end{document}